\documentclass[preprint,journal]{vgtc}       

\ifpdf
  \pdfoutput=1\relax                   
  \usepackage{graphicx}                
  \DeclareGraphicsExtensions{.pdf,.png,.jpg,.jpeg} 
\else
  \usepackage{graphicx}                
  \DeclareGraphicsExtensions{.eps}     
\fi%

\graphicspath{{figures/}{pictures/}{images/}{./}} 

\usepackage{microtype}                 
\PassOptionsToPackage{warn}{textcomp}  
\usepackage{textcomp}                  
\usepackage{mathptmx}                  
\usepackage{times}                     
\usepackage{cite}                      
\usepackage{tabu}                      
\usepackage{booktabs}                  
\usepackage{amsmath}

\usepackage{stfloats}

\newcommand{\change}[1]{\textcolor{black}{#1}}

\usepackage[normalem]{ulem}
\newcommand{\remove}[1]{}

\onlineid{xxxx}

\vgtccategory{Research}
\vgtcpapertype{Algorithm/Technique}

\title{VirtualCube: An Immersive 3D Video Communication System}

\author{Yizhong Zhang*, Jiaolong Yang*, Zhen Liu, Ruicheng Wang, Guojun Chen, Xin Tong, 
and Baining Guo,  \\ \textit{Fellow, IEEE}}
\authorfooter{
\item
 Yizhong Zhang, Jiaolong Yang, Xin Tong, Guojun Chen, and Baining Guo are with Microsoft Research Asia, Beijing, China. \\
 Email: \{yizzhan, jiaoyan, guoch, xtong, baingguo\}@microsoft.com
\item
 Zhen Liu is with Nanjing University, Nanjing, China.  Work done during internship at Microsoft Research Asia. Email: zhenliu@smail.nju.edu.cn
\item
 Ruicheng Wang is with University of Science and Technology, Hefei, China. Work done during internship at Microsoft Research Asia.  \\ 
 Email: wangrc2018cs@mail.ustc.edu.cn
\item
 * Joint first authors with equal contribution; order determined by coin flip.
 
}

\abstract{
The VirtualCube system is a 3D video conference system that attempts to overcome some limitations of conventional technologies. The key ingredient is VirtualCube, an abstract representation of a real-world cubicle instrumented with RGBD cameras for capturing the user’s 3D geometry and texture. We design VirtualCube so that the task of data capturing is standardized and significantly simplified, and everything can be built using off-the-shelf hardware. We use VirtualCubes as the basic building blocks of a virtual conferencing environment, and we provide each VirtualCube user with a surrounding display showing life-size videos of remote participants. To achieve real-time rendering of remote participants, we develop the V-Cube View algorithm, which uses multi-view stereo for more accurate depth estimation and Lumi-Net rendering for better rendering quality. The VirtualCube system correctly preserves the mutual eye gaze between participants, allowing them to establish eye contact and be aware of who is visually paying attention to them. The system also allows a participant to have side discussions with remote participants as if they were in the same room. Finally, the system sheds lights on how to support the shared space of work items (e.g., documents and applications) and track participants’ visual attention to work items.
} 

\keywords{3D video, Teleportation, Telecollaboration}

\teaser{
  \centering
  \includegraphics[width=\linewidth]{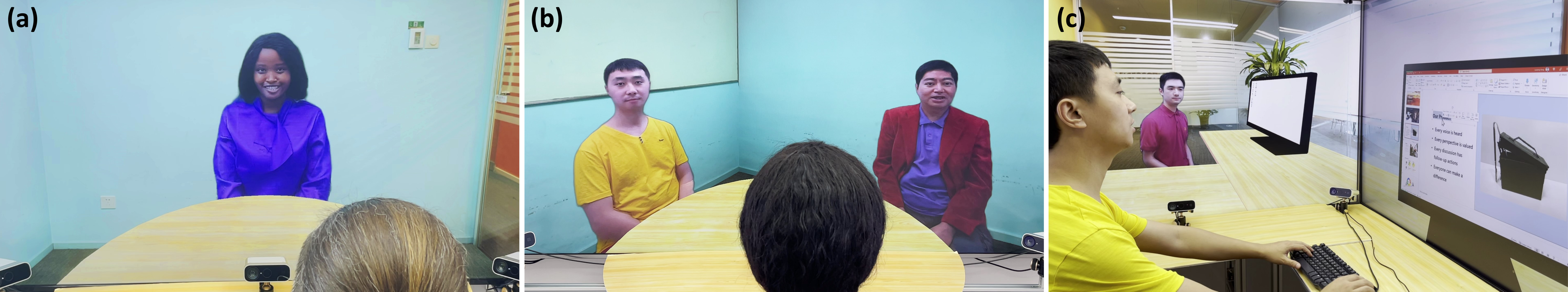}
  \vspace{-18pt}
  \caption{Snapshots of the VirtualCube system in action, with the local participant in the foreground. The images of remote participants on the screen are synthesized from the RGBD data acquired by cameras.  (a) A face-to-face meeting with two participants.  (b) A round-table meeting with multiple participants, each in a different location. No two participants are in the same location. (c) A side-by-side meeting that includes sharing work items on the participants’ screens, as if the participants were sitting next to each other working together. Our system achieves mutual eye contact and visual attention as in in-person meetings. The lively recorded videos can be found on the  \href{https://www.microsoft.com/en-us/research/project/virtualcube/}{\uline{{project page}}}.
  }
	\label{fig:teaser}
}

\vgtcinsertpkg

\begin{document}
\maketitle
\section{Introduction}
In a real-life conversation, many visual cues, eye contact, and gaze direction 
contribute substantially to
communication and thus make the conversation much more effective than video conferencing \cite{buxton1997,Chen2002}. Nevertheless, video conferences have become increasingly important because remote collaboration over long distances -- either for work or for entertainment -- has become commonplace. It is often simply too time-consuming and costly to bring remote parties to the same location. Furthermore, the recent COVID pandemic has taught us that sometimes it is impossible to bring people together physically and it has made a distributed workforce a permanent part of many organizations worldwide. Thus, it is of significant interest to develop video conferencing technologies that facilitate more effective communication \cite{Jones2009,kuster2012gaze,elgharib2020egocentric} and possibly create the illusion that the remote participants are in the same room \cite{Raskar98,Gross2003,Matusik2004,Beck2013,Orts2016}. 

Conventional video conference technologies unfortunately fall short of expectations. For example, with most existing technologies it is difficult to establish eye contact with remote participants and be aware of who is paying attention to you visually. Yet head turning and mutual eye gaze is such an essential part of everyday conversation and is commonly used to manage turn-taking and floor control in in-person  meetings. Another limitation of existing technologies is that it is difficult to 
glance sideways at and have side conversations with 
other participants, which are also common practices in in-person meetings.

In this work, we present the VirtualCube system that attempts to overcome some of these limitations. The key ingredient of the system is VirtualCube (or V-Cube for short), which is an abstract representation of a standardized real-world cubicle instrumented with multiple RGBD cameras for acquiring the user’s 3D geometry and texture. We only capture the user’s upper body so that we not only obtain the most important visual gestures, including facial and hand gestures, but also protect the user’s privacy to some extent. VirtualCube provides the user with a surrounding display showing life-size videos of remote participants. This surrounding display consists of three large-format screens on the front and two side walls around the user. We also segment the user from the background. Thus VirtualCube essentially represents the user’s 3D geometry and texture and the enclosing geometric cube corresponding to the physical cubicle. 

An important feature of VirtualCube is that it is easy to set up most video conferences by using VirtualCubes as basic building blocks. As a result, the VirtualCube system provides a highly  versatile way to conduct video conferences. With VirtualCube, we can build video conferences for the common scenarios of face-to-face meetings with two participants and round table meetings with multiple participants. We can also model the important scenario of a side-by-side meeting in which both participants and their desktops are all part of the meeting, as if the participants were sitting next to each other working together. This last scenario is quite common in real life but receives only limited support with existing video conferencing technologies. When a video conference is constructed from a set of adjacent VirtualCubes sitting on a common floor, we call the resulting assembly of VirtualCubes a V-Cube Assembly. In essence, we can view the V-Cube Assembly as the global virtual environment of the video conference. 

The VirtualCube system provides each meeting participant with real-time rendering of all remote participants on his/her surrounding display, thanks to 
the acquired 3D geometry and texture data of all constituent VirtualCubes of the video conference. Most importantly, the rendering correctly preserves the mutual eye gaze between participants because the VirtualCube system renders the remote participants in life-size using their positions in the global virtual environment and the local participant’s view position in the global virtual environment, just as if the local participant was seeing the remote participants through the screens of the VirtualCube. The VirtualCube system can perform such rendering since it knows both the 3D geometry of all participants and the geometry of their VirtualCubes (both their dimensions and absolute positions in the virtual environment) and thus can properly carry out all the 3D geometric transformations needed for correct rendering.  

For rendering remote participants, real-time frame rates are essential. To achieve real-time performance while improving rendering quality, we develop the V-Cube View algorithm for novel view synthesis. For a given viewpoint, we adapt multi-view stereo to more accurately estimate the depth of each image pixel. To best compute the pixel color from the acquired textures, we develop the Lumi-Net rendering technique for geometry-aware texture blending based on lightfield/lumigraph principles~\cite{levoy1996light,gortler1996lumigraph,debevec1996modeling,buehler2001unstructured}. 
Both the depth estimation and Lumi-Net rendering are carefully crafted to ensure that the overall rendering process achieves real-time performance.

In summary, the VirtualCube system has the following advantages:
\begin{itemize}
    \item Standardized and simplified, all using off-the-shelf hardware. 
    As every VirtualCube is made the same, 
    the VirtualCube system provides a consistent physical environment and device setup which not only simplifies the workload of device calibration but also reduces the difficulty of 3D video capturing and processing. Also, we only capture the upper body of a seated user, which greatly reduces the workload for capturing and modeling of complex human poses. \remove{Finally, VirtualCube can be built completely with off-the-shelf hardware.}
    \item Versatile modeling. A set of VirtualCubes can be easily assembled into a V-Cube Assembly to model different video communication scenarios, including scenarios which are poorly supported by existing technologies.
    \item Real-time, high-quality rendering. The real-time rendering can capture a variety of subtle surface appearances, such as glossy reflection on human faces and clothing, as shown in the accompanying video.  
\end{itemize}

Our experiments with the VirtualCube system show that it can bring video conferencing significantly closer to our everyday experience of in-person meetings. The VirtualCube system allows meeting participants to establish eye contact and be aware of who is visually paying attention to them. The system also allow a participant to 
make side glances and side conversations to 
remote participants as if they are in the same room. Taken together, these abilities form the cornerstone of the notion of a personal space, a notion that we have acquired and internalized through our lifetime experience of in-person conversations~\cite{buxton1997}. 

In addition to this shared “person space”, the VirtualCube system also shows promise in supporting the shared space of work items (including documents, applications, and other artifacts). In particular, in a side-by-side meeting, both participants and their desktops are all part of the video conference, as if the two participants were sitting next to each other working together. This shared “work space” \cite{buxton1997}, which encompasses both the participants and work items on their desktops, is a useful addition to the shared “person space” because it lets each participant see whether the remote participant is visually attending to a specific work item as desired. For example, when a participant highlights an item on his/her desktop with the cursor, he/she can easily see whether the remote participant is visually paying attention to this item. This experience, albeit simple and very common for in-person collaboration \cite{buxton1997}, is not supported in conventional video conferences.

\section{Related Work}
The techniques and systems for 3D video communication have been widely studied in many fields. In this section, we discuss the previous works that are directly related to our work, including 3D video communication systems, gaze-correction techniques for video conferencing, and free viewpoint video of human characters. 
For more comprehensive reviews of related techniques, the reader is referred to the latest works \cite{wang2021facevid2vid,shi2020learning,Tausif2020} and surveys \cite{nagata2017virtual,ohl2018tele}.

\subsection{3D Video Communication}
Early 3D video communication systems \cite{gibbs1999teleport,baker2002coliseum,sadagic2001tele} attempted to capture the appearance and geometry of the participants at different locations and place them in a common virtual environment. Without careful system design and setup, the mutual gaze contact 
among users in these systems is always poorly produced. 

Many follow-up techniques~\cite{nguyen2005multiview,kuechler2006,Jones2009,Pluss2016,kuster2012,gotsch2018telehuman2} have been developed for telecommunication between two sites. Although these approaches improved eye contact and user experience in two-site meetings, extending them to multiple sites is difficult.  

For three-party teleconferencing, the tele-cubicle systems~\cite{wen2000,Towles2002} use a cubicle with two walls. In their systems, users are difficult to establish eye contact~\cite{Towles2002}. Their focus is to use the two walls as portals to remote users’ offices; it is not their intention to bring remote users into a common virtual meeting room as we do. Also note that the remote users are reconstructed as 3D meshes in their system. Real-time reconstruction of detailed 3D meshes that permit high-quality rendering cannot be easily achieved even with modern hardware and algorithms~\cite{Orts2016,collet2015high}.
Later, a number of systems \cite{kauff2002immersive,Zhang2013} have been proposed where the participant positions in the virtual environment are predefined and fixed for maintaining mutual eye contact. 
However, it is unclear how to extend these systems to other seating arrangements and meetings with different numbers of participants. Also, the rendering quality of the remote participants is limited due to the fragile depth reconstruction and image warping algorithms used in these systems. In contrast, our system is designed for implementing meetings between different numbers of participants, as well as meetings with various user seating setups. To maintain mutual eye contact of the participants in different meeting setups, we develop a new deep learning based V-Cube View algorithm for capturing and rendering high-fidelity free-viewpoint video of the participants in real time.

A collection of telepresence systems~\cite{Raskar98,Gross2003,Matusik2004,Beck2013,Fuchs2014} have been designed for capturing and rendering a virtual or augmented-reality environment so that the participants at different sites could work together as if they were in one site. All these methods focus either on the environment capturing and 3D display~\cite{Raskar98,Gross2003,Matusik2004,Fuchs2014} or the interactions between the users and the virtual environment~\cite{Beck2013}. They are not optimized for teleconference and thus it is difficult for them to achieve faithful mutual gaze contacts between the participants at different sites.   

\subsection{Gaze Correction for Video Conferencing}
Many approaches have been developed to correct the gaze of the participants for better eye contact in video conferencing. Chen~\cite{Chen2002} studied the sensitivity of eye contact in video conferencing and improved the eye contact with more accurate horizontal gaze direction. Kuster et al.~\cite{kuster2012gaze} warped the face region of the remote participant captured by an RGBD camera and then fused the warped part into the original RGBD video frame for correcting vertical gaze direction. Later, Giger et al.~\cite{Giger2014} extended this method for RGB videos captured by a web camera. 
Hsu et al.~\cite{Hsu2019} developed a convolutional neural network to generate gaze corrected video of the participant by tracking and warping the eye region of the participant in each video frame.  
Tausif et al.~\cite{Tausif2020} moved a web camera behind a transparent screen according to the remote participant's eye position for capturing videos of the participants with correct eye-contact. All these methods are designed for video conferencing between two participants, and cannot handle eye contact for three or more participants.    

Different from these approaches, our method captures 3D video of the participants and delivers natural gaze contact and other visual communications by realistically rendering  life-size remote participants from the local participant's viewpoint. Our method supports eye contact between the participants in a teleconference with two or more users.    

\subsection{Free Viewpoint Video of Human Characters}

Numerous methods have been proposed for generating a realistic avatar of a subject. Some of them~\cite{Oleg2009,Kyle2016,Thies2018,Wei2019,Lombardi2019,Thies_2016_CVPR,thies2015realtime,Tewari2019,Jiahao2018,Koki2018,Kyle2017,kim2018deep,Timur2021TOGFullbodyavatar} model a 3D avatar of a subject and then drive its animation using a video sequence or speech text. However, dedicated device setup and heavily manual work are always needed for generating a realistic avatar and reconstructing the detailed appearance, subtle expressions, and gaze movement of a subject. Recent deep-learning based methods \cite{nirkin2019fsgan,Wangvid2vid2019,wang2021facevid2vid,zakharov2020fast,xu2020deep,zakharov2019few,zhou2019talking,burkov2020neural,chen2019hierarchical,gu2020flnet,ha2020marionette} avoid 3D avatar modeling and directly synthesize a talking head video of a subject from one source image of the subject and a video sequence. Elgharib et al.~\cite{elgharib2020egocentric} developed a solution for warping the video of a subject's face from side view to front view. 
Unfortunately, all these deep-learning based methods cannot support arbitrary view rendering. Also, it is unclear how to extend these methods for modeling the dynamics of the upper body.  

Other methods ~\cite{kauff2002immersive,maimone2011encumbrance, kuster2011freecam,Zhang2013,zollhofer2014real,Orts2016,dou2016fusion4d,Guo2017} reconstruct a complete or partial 3D geometry and texture of a dynamic character in real time and then render it from novel viewpoints. Although they provide real-time free-viewpoint rendering of the dynamic character, the reconstructed 3D geometry and texture 
are always imperfect or of low-resolution thus leading to inferior rendering quality.

Generic image-based rendering methods have been developed for synthesizing novel views from images or videos of a scene captured from sparse or dense views. Traditional optimization-based methods \cite{collet2015high,zitnick2004high,lipski2010virtual,chaurasia2013depth, penner2017soft} rely on fragile offline processing for obtaining high quality rendering results. Deep-learning based approaches leverage neural networks to improve the robustness and speed of offline optimization and have demonstrated high-quality novel view synthesis results~\cite{kalantari2016learning,shi2020learning, shi2020learning,flynn2016deepstereo,choi2019extreme,zhou2018stereo,flynn2019deepview}. However, these methods still require at least a few seconds on a modern GPU to synthesize a novel view. Recently, several neural representations~\cite{mildenhall2020nerf, yu2021pixelnerf,wang2021ibrnet,chen2021mvsnerf} have been proposed for modeling and rendering a 3D scene from a given multiview image collection. Unfortunately, the computational cost of learning and rendering these neural representations is still high. Although these image-based approaches can be applied for rendering human characters, they suffer from the trade-off between speed and quality. To the best of our knowledge, methods have not been developed for 
online capturing and realistic rendering in real-time. 

In this work, we follow the principle of previous deep-learning based methods \cite{kalantari2016learning,shi2020learning,flynn2016deepstereo,choi2019extreme} that first predicts the target-view depth and then synthesizes texture, and develop a new \emph{V-Cube View} method to achieve real-time and high-quality 3D freeview synthesis. 

\section{The VirtualCube System}
\label{sec:system}
In this section, we first provide a conceptual overview of VirtualCube and then present the implementation details. Our real-time rendering algorithm is described in Section \ref{sec:3dvideo}.
 
\subsection{VirtualCube: A Conceptual Overview}

\begin{figure}
    \centering
    \includegraphics[width=\columnwidth]{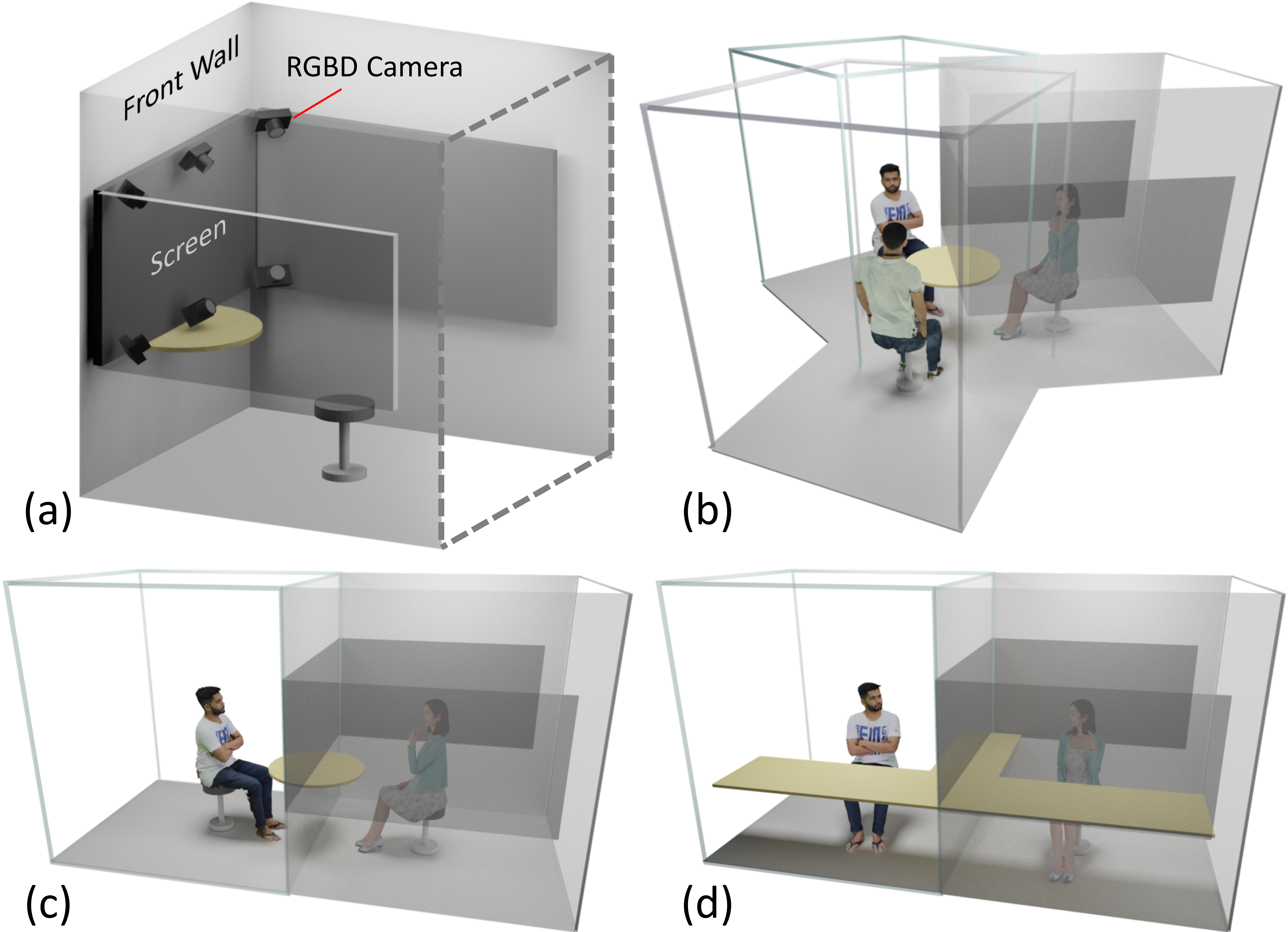}
    \caption{VirtualCube and various examples of video conferences constructed by using VirtualCubes as building blocks. (a) The physical setup of VirtualCube, which consists of a surrounding display on the front and two side walls and multiple RGBD cameras mounted around the screen on the front wall. The back wall is outlined by dotted lines. (b) A round table meeting of three participants. (c) A face-to-face meeting of two participants. (d) A side-by-side meeting of two participants. To simplify illustration, we only show one participant’s display in (b), (c), and (d).
    }
    \label{fig:system_hardware}
    \vspace{-3pt}
\end{figure}
 
The VirtualCube system aims to provide 
a standardized solution
for remote meeting participants to naturally communicate with one another as if they were in the same room. For simplicity, we assume that no two participants are co-located. The immersive experience of a participant is created by surrounding him with a large format display on which all other participants are rendered at life size. Meanwhile, at each meeting site, multiple cameras capture the local participant’s 3D geometry and texture, and thus allow our system to synthesize the video of this participant for all remote parties. 
 
VirtualCube has two components: one physical and the other abstract. The physical component is an instrumented real-world cubicle enclosed by four vertical walls as shown in Fig.~\ref{fig:system_hardware}(a). The cubicle has a fixed seat in the middle for the user and a table attached to the front wall. The front and two side walls have large format screens attached for displaying life-size videos of remote meeting participants. Along the screen boundaries of the front wall a set of RGBD cameras (color and depth) are attached for capturing the user’s 3D geometry and texture. The back wall is painted solid gray for simplifying of the task of segmenting the user from the background. The abstract component of VirtualCube consists of the abstract geometric cube associated with the physical cubicle and the user’s 3D geometry and texture as captured by the RGBD cameras. 
 
The VirtualCube system only captures and renders the 3D geometry and color texture of the seated user’s upper body, defined as the body part above the table. Our reason for focusing on upper body is mainly for efficiency. The upper body of a seated person is easier to capture than the full body with unconstrained movements. Yet, it provides facial and hand gestures which are among the most important visual cues for communication. In practice, only rendering the upper body also provides a sense of privacy for the user who knows the lower body is safely off the limits and hence any lower garment can be worn.
 
We use VirtualCube as the basic building block of the virtual environment of our online meetings. This virtual environment is constructed as an assembly of multiple adjacent VirtualCubes on a common floor. We call this assembly the \emph{V-Cube Assembly}. Figure~\ref{fig:system_hardware}(b)-(d) show a few V-Cube Assemblies in different configurations: one for a face-to-face meeting with two participants, one for a side-by-side meeting with two participants, and one for a round table meeting with multiple participants. Note that the side-by-side configuration has a special advantage of supporting not only the interaction between the participants but also the sharing of their screen content as part of the interaction.

People have different senses of personal spaces. We support this by allowing different VirtualCubes in a V-Cube Assembly to overlap. Intuitively, the overlap allows participants to sit more closely together when desired. The amount of overlap is a user-controlled parameter, which can be set by the conference organizer. As a rule of thumb, in our implementation we require the meeting participant in each VirtualCube to be outside of the other VirtualCubes in the V-Cube Assembly 
so that we can correctly project the remote participants on the surrounding display 
from the view of each participant.

The V-Cube Assembly is the basis of the global coordinate system, which is defined as the coordinate system of the V-Cube Assembly and hence the overall virtual environment. This global coordinate system is in contrast to the local coordinate system of each individual VirtualCube. The correct 3D geometric transformations between the global and local coordinate systems (which take into consideration the physical size of VirtualCube and its display screens) is important because it is essential for correct rendering of remote participants on the video display of each meeting participant. In particular, this correct rendering is needed for achieving mutual eye gaze. Specifically, the 3D geometry transformations proceed as follows. First, when capturing the user’s 3D geometry and texture inside a VirtualCube, we use the VirtualCube’s local coordinate system. Then, after forming the V-Cube Assembly, we transform the captured 3D geometry data of all constituent VirtualCubes from their respective local coordinate systems into the global coordinate system so that all VirtualCubes’ 3D geometry data are correctly positioned in the global virtual environment to create the global 3D geometry content of the V-Cube Assembly. This global 3D geometry is what every participant sees. Finally, to render remote participants on a VirtualCube’s display, we first transform the global 3D geometry into the VirtualCube’s local coordinate system and then project the transformed global geometry onto the VitualCube’s display. 

\subsection{Implementation Details}
We now present our hardware setup and calibration and the resulting coordinate transformations for the video display of each VirtualCube.  

\begin{figure}
    \centering
    \includegraphics[width=\columnwidth]{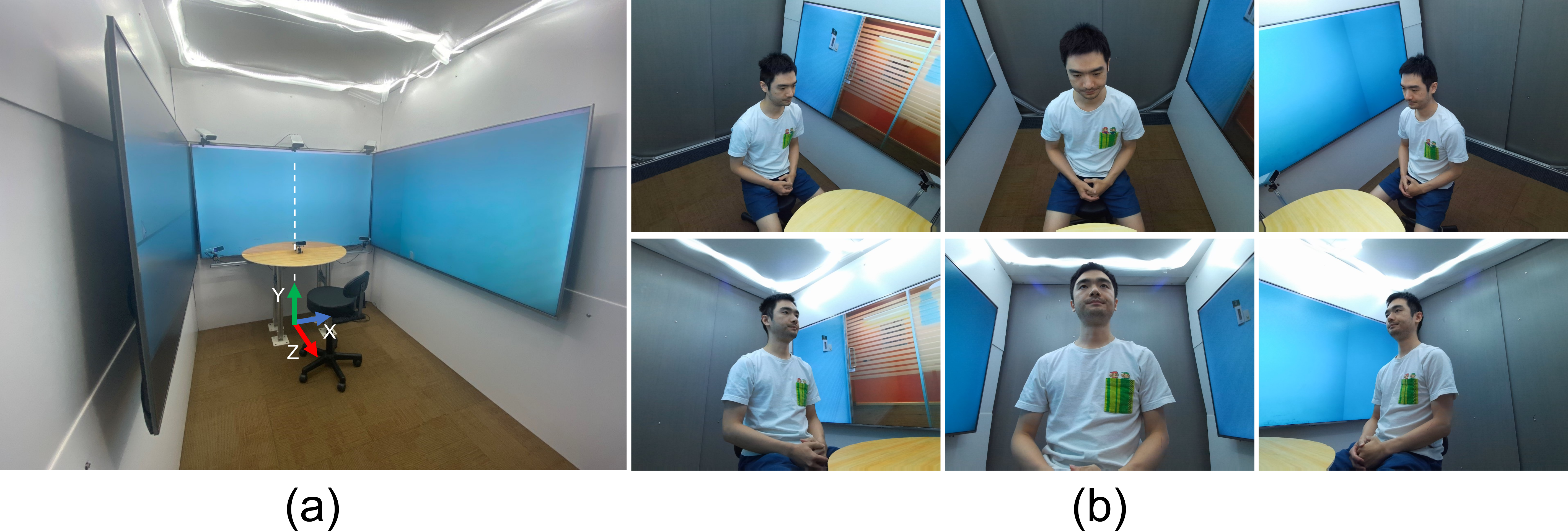}
    \vspace{-15pt}
    \caption{(a) The hardware setup of our VirtualCube implementation, which consists of three large screens on the front and two side walls and six Azure Kinect cameras mounted around the front display. We also show the VirtualCube’s local coordinate system. (b) Color images of the participant captured by the six cameras.}
    \label{fig:physical_cube}
\end{figure}
 
\paragraph{Hardware setup} As shown in Figure \ref{fig:system_hardware}, our VirtualCube prototype is a cubicle enclosed by the front and two side walls painted solid white, and a back wall covered by a curtain of solid grey color for easy segmentation of the user from the background. The floor plane dimension of VirtualCube is $1.6 \times 2.0$ meters. 
In face-to-face meetings and round-table meetings, the user faces the front display and a semi-circular table is placed between the user and the front wall. In side-by-side meetings, the user faces a side display and an L-shaped long table is placed between the user and the front wall and one side wall, as shown in Fig.~\ref{fig:system_hardware}(d).

For the surrounding display of a VirtualCube, we mount three 65-inch 4K flat LCD screens on the front and two side walls. The height from the bottoms of these screens to the floor is $0.7$ meter, which is designed to support life-size display of the upper bodies of remote participants. For capturing the 3D video of the user, we install six AzureKinect RGBD cameras around the front screen, with four at the corners and two at the mid-points of the upper and lower boundaries. The viewing directions of the six cameras are set towards the user seated in the center of the VirtualCube. The six cameras capture RGBD video sequences in synchronized mode, each recording $2560 \times 1440$ RGB frames and $640 \times 576$ depth frames at 30fps. The RGB and depth frames are aligned and scaled to $1280\times 960$ resolution for later use.
 
Within VirtualCube, the front and two side screens and six RGBD cameras are connected to a PC workstation with GPUs for 3D video capturing, rendering, and display. VirtualCube instances at different sites are connected by a network.

\paragraph{Hardware calibration} For the hardware calibration of a VirtualCube, we need to define the local coordinate system of the VirtualCube and calibrate the intrinsic/extrinsic parameters of all six RGBD cameras. We also need to compute the positions and orientations of the front and side screens in the VirtualCube’s local coordinate system. 
 
We start by defining the VirtualCube’s local coordinate system. As shown in Fig.~\ref{fig:physical_cube}(a), we carefully adjust the orientation of the screen on the front wall so that the horizontal direction of the screen (i.e., the direction of the pixel rows) is parallel to the floor and the screen plane is perpendicular to the floor. Based on this setup, we define the $X$ direction of the local coordinate system as the horizontal direction of the front screen and the $Y$ direction as the upwards direction that is perpendicular to the floor plane. The $Z$ direction is determined by the cross product of the $X$ and $Y$ directions accordingly. The $XZ$ plane is the floor plane, and the origin of the local coordinate system is defined as the intersection point between the central vertical line of the front screen and the floor plane. The scale of this local coordinate system is set to be the same as the scale of the physical world. 

Next we calibrate the positions and orientations of the three display screens in the local coordinate system. We measure size of each screen and adjust the the two side screens so that they are perpendicular to the front screen with their side edges seamlessly aligned with the side edges of the front screen. In this way, the two side screens are parallel to the $YZ$ plane and their positions can be easily computed.

For camera calibration, we use the intrinsic parameters of RGB and depth cameras provided by AzureKinect SDK and calibrate the extrinsic parameters (i.e., camera poses). We first calibrate the poses of the six cameras in a local coordinate system using the camera calibration toolkit in OpenCV and a checkerboard pattern. 
To transform the camera poses from this local coordinate system to the VirtualCube’s local coordinate system, we manually measure the distance of the camera to the surface and boundaries of the front screen and use these distances to calculate the transformation.

\paragraph{View positioning}
To compute the viewpoint of the participant in a VirtualCube, a real-time viewpoint tracker is implemented. We define the participant's viewpoint as the middle point of the line segment joining the eye centers. To obtain the viewpoint for each frame, we first detect the 2D eye centers on the multiview RGB images using a fast facial landmark detector \cite{chen2014joint} and then lift them to 3D using linear triangulation~\cite{hartley2003multiple}. Images with no landmark detected due to occlusion are simply discarded.
With six cameras installed in a VirtualCube, the implemented method can robustly track the participant's 3D viewpoint in real-time during video communication.  

\begin{figure}
    \centering
    \includegraphics[width=\columnwidth]{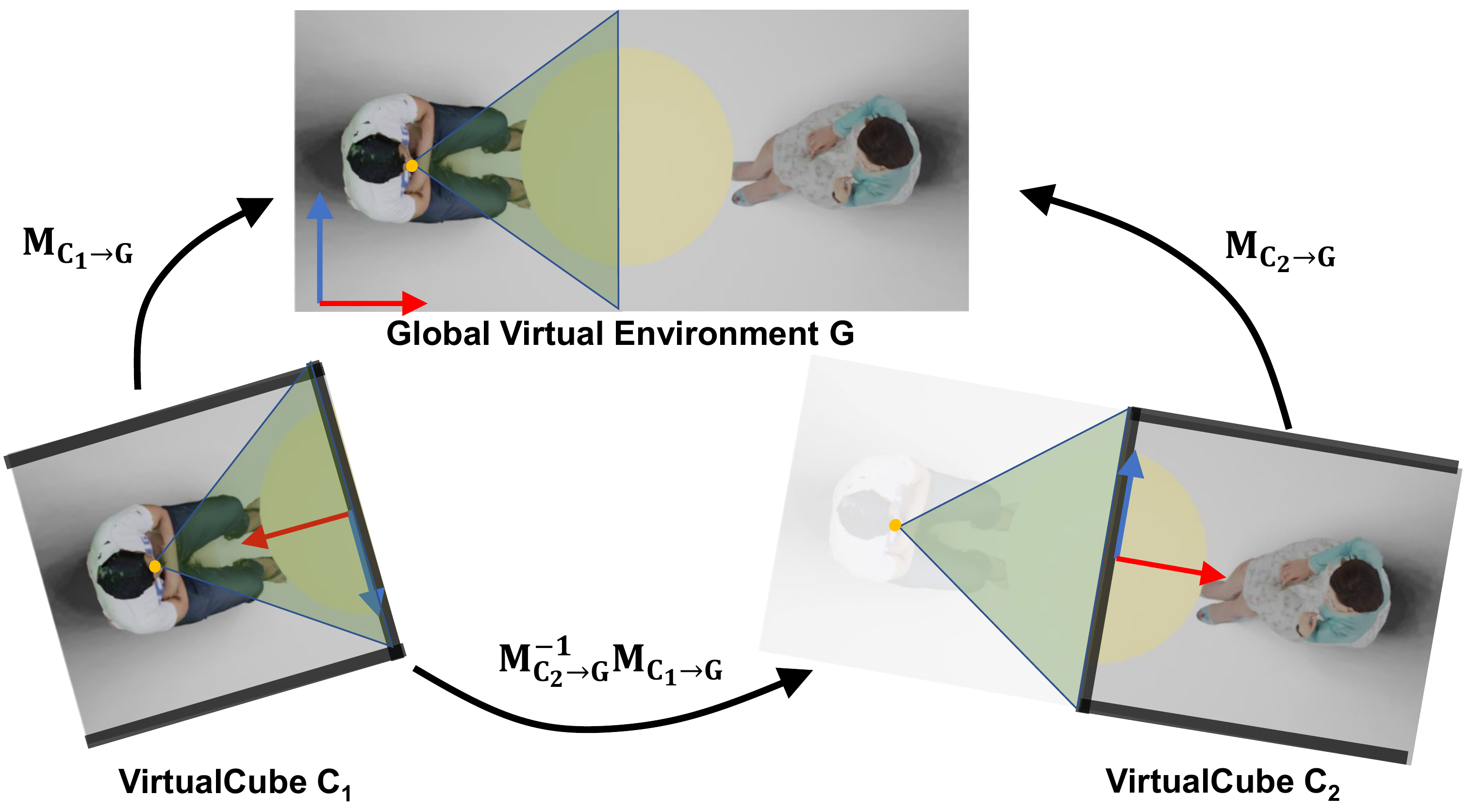}
    \caption{
    Coordinate transformations in the VirtrualCube system.
    Conceptually we map every constituent VirtualCube into the global coordinate system defined by the V-Cube assembly, which is the global virtual environment. In the above example, the global virtual environment consists of two VirtualCubes. In practice, we transform the viewpoint of the receiver site participant (orange point in VirtualCube $C_1$) into the local coordinate system of the sender site VirtualCube $C_2$ for rendering the participant in $C_2$ on the display in VirtualCube $C_1$.}
    \label{fig:transformation}
\end{figure}

\paragraph{Transformation to global coordinate system} When we assemble a set of VirtualCubes into a V-Cube Assembly, we define both a 3D virtual environment and a global coordinate system. Once this global virtual environment is defined, we need to transform every constituent VirtualCube into the global coordinate system. To do this, we first obtain the position and orientation of each VirtualCube $C_i$ in the global coordinate system and then compute the transformation $M_{C_i \rightarrow G}$ that maps the VirtualCube $C_i$ into the global virtual environment, where $G$ refers to the global virtual environment defined by the V-Cube Assembly (as shown in Fig.~\ref{fig:transformation}). 
We assume the scale of the global virtual environment to be the same as the scale of the physical world.
With this transformation, we map all local entities defined in the local coordinate system of VirtualCube $C_i$, including the user’s 3D geometry and texture, the surrounding display screens, and the user’s viewpoint, into the global coordinate system. Now for each meeting participant, we can render the remote participants and the background virtual environment with the view frustums determined by this participant’s viewpoint and three display screens in the global coordinate system. For the participant in VirtualCube $C_i$, the life-size image so rendered provides the correct positions, orientations, sizes, and perspectives of the remote participants and thus gives rise to the illusion that they are in the same room. In this way, the mutual eye gaze between participants is naturally preserved.

\begin{figure}[t]
	\centering
	\includegraphics[width=\columnwidth]{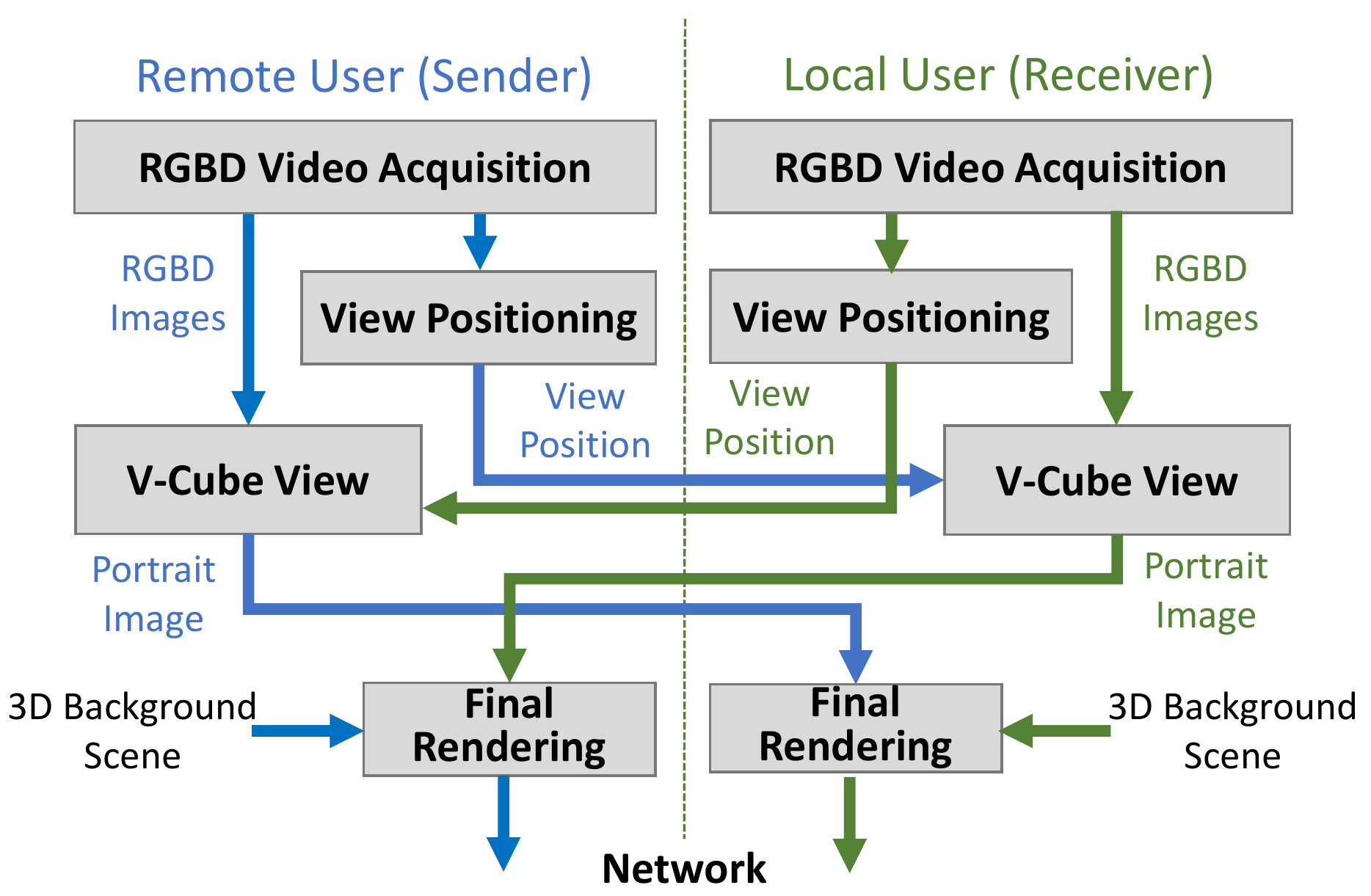}
	\caption{Data and workflow of the VirtualCube system, where the sender's portrait image is rendered from the receiver's view position and then transmitted to the receiver side to composite with the rendered 3D scene for final display. The portrait image includes RGB and alpha channels to facilitate composition in the final rendering.}
	\label{fig:system_software}
\end{figure}
 
\paragraph{Workflow of the VirtualCube system} As shown in Fig.~\ref{fig:system_software}, every VirtualCube is simultaneously a sender and receiver of rendered images. Every VirtualCube also knows the global coordinate system of the V-Cube Assembly, the local coordinate systems of all constituent VirtualCubes, and the viewpoints of all meeting participants. At the sender site, our V-Cube View rendering algorithm generates the portrait image and alpha mask of the sender site’s participant using the receiver site’s viewpoint and sends the rendered RGBA image to the receiver site. During rendering, proper coordinate transformations are performed to ensure correct rendering and mutual eye gaze: the sender VirtualCube’s 3D geometry content is first transformed into the global coordinate system for correct positioning in the global virtual environment and then transformed into the receiver VirtualCube’s local coordinate system for rendering. 

In practice, we render the image of the sender site participant in VirtualCube $C_i$ for the receiver site VirtualCube $C_j$ by transforming the viewpoint of the receiver site participant into the local coordinate system of the sender site VirtualCube $C_i$ via a transformation $M^{-1}_{C_i\rightarrow G}M_{C_j \rightarrow G}$. 
We then use the RGBD frames of four cameras that are close to the transformed viewpoint for rendering the image of the sender site participant. 
To this end, our V-Cube View algorithm renders the participant from the segmented RGBD frames and sends the rendered color image and alpha mask to receiver VirtualCube $C_j$ (as shown in Fig.~\ref{fig:transformation}). After receiving the portrait RGBA images from all other VirtualCubes, the receiver site generates its final rendering display by compositing  all of the incoming portrait images against a common background, which is the rendering of a 3D virtual conference room enclosing the V-Cube Assembly. 
 
The above work flow is designed to improve the rendering quality of the sender’s video. Conceptually, an alternative work flow would be to send all VirtualCubes’ 3D geometry and texture content to the receiver site and assemble the global 3D geometry of the entire virtual environment and then render this global 3D geometry. However, when a global 3D geometry model is reconstructed at the receiver site, the common practice is to build a view-independent 3D geometry model and this reconstruction often leads to the loss of geometry details. We do not perform such a 3D reconstruction, but instead we only perform the depth estimation for the receiver site’s viewpoint using the raw depth data acquired at the sender site. By doing so we maximize the utilization of the captured raw 3D geometry data, minimize unnecessary geometry detail loss usually associated with a full 3D geometry reconstruction, and thus improve the rendering quality.  

\paragraph{V-Cube View input segmentation} 
Our system segments the participant from the input frames and feeds the segmented images into the V-Cube View algorithm. Different from the conventional video conferencing where the segmented frames are directly used for display, our algorithm takes the segmented input for improving the speed and robustness of novel view synthesis. So the segmentation results need not be accurate. For this purpose, we roughly segment the foreground region (i.e., the portrait of the participant) from each input frame by a simple algorithm based on background color and depth comparison. Specifically, we capture background RGBD images from six cameras before the user enters the cubicle. At run-time, we compute the differences of pixels' depth and color to the background image values and subtract foreground pixels whose depth or color difference is larger than a pre-defined threshold (10cm/30 gray level in our implementation).

\newcommand{\bI}{\mathbf{I}}
\newcommand{\bD}{\mathbf{D}}
\newcommand{\bM}{\mathbf{M}}
\newcommand{\bW}{\mathbf{W}}
\newcommand{\bF}{\mathbf{F}}
\newcommand{\bV}{\mathbf{V}}
\newcommand{\bP}{\mathbf{P}}
\newcommand{\bN}{\mathbf{N}}
\newcommand{\overbar}[1]{\mkern 1.5mu\overline{\mkern-1.5mu#1\mkern-1.5mu}\mkern 1.5mu}
\begin{figure*}[t]
	\centering
	\includegraphics[width=0.988\linewidth]{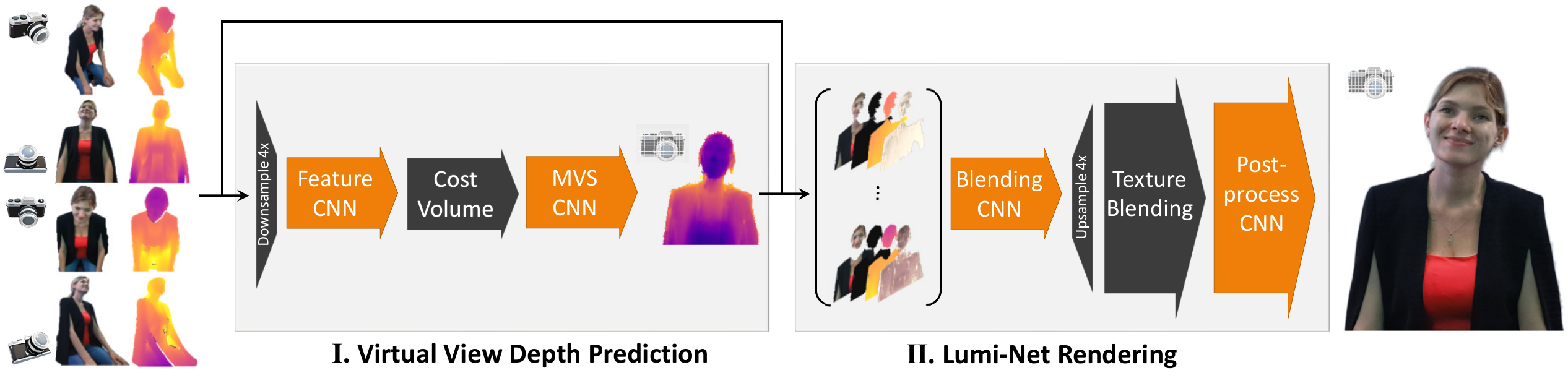}
	\vspace{-3pt}
	\caption{Overview of the V-Cube View algorithm. The inputs are four $1280\times 960$ RGBD images and a virtual viewpoint. The output is a $1280\times 960$ RGB$\alpha$ image at the virtual viewpoint synthesized in real time on a modern GPU card.
	}
	\vspace{0pt}
	\label{fig:3dvideo_pipeline}
\end{figure*}

\section{V-Cube View for Real-time Rendering}\label{sec:3dvideo}
When designing a rendering algorithm for VirtualCube we face three challenging issues. The first is real-time performance. The VirtualCube acquires and segments RGBD images in real time. The rendering algorithm must also run at real-time rates for the whole VirtualCube system to function well. The second is high rendering quality. Our VirtualCube displays life-size human portrait videos on large screens where rendering flaws can be easily noticed by users.
However, we are not aware of any novel view synthesis method that can perform high quality rendering in real time. Most existing methods cannot achieve online capturing and rendering, while other real-time solutions suffer from severe artifacts, producing incomplete regions or only generating low-resolution mesh textures.
Finally, we have to deal with wide baselines in rendering. Due to our large screen sizes, the view difference between the cameras mounted around the screen are large and the rendering must cover a wide range of virtual viewpoints. 

To address these issues we develop the V-Cube View algorithm based on a few key insights. First, we leverage depth cameras to ease the burden of virtual view geometry estimation. The acquired depth maps, although quite noisy and cannot be directly used, can provide reasonable depth initializations of the virtual view for a lightweight refinement process. Second, we observe that several computation intensive procedures can run at low resolutions to save cost. Thus the main body of our algorithm, including depth estimation and texture blending weight prediction, runs at low image resolution (quarter size), after which we blend the warped textures and apply a lightweight post-processing CNN at the full resolution. Third, we propose a novel Lumi-Net rendering weight prediction scheme. Inspired by traditional image-based rendering techniques, we incorporate view direction and depth difference priors to predict geometry-aware blending weight maps, which lead to improved rendering quality especially for our wide-baseline scenario. Finally, we observe that the fidelity of human faces are critical in video conferencing, as people are much more attuned to perceiving facial details than other content. Thus we introduce a perceptual loss on the face region during training, which significantly improved the face quality in rendering.

Figure~\ref{fig:3dvideo_pipeline} provides an overview of the V-Cube View algorithm. The input are 4 background-subtracted RGBD images of $1280\times 960$ resolution and their associated camera poses, and a target virtual camera pose. The input camera poses are readily available because all cameras in VirtualCube are pre-calibrated to obtain the intrinsic and extrinsic parameters. The output is a $1280\times 960$ RGB$\alpha$ portrait image for the given virtual viewpoint. We only activate 4 cameras out of the 6 in the VirtualCube to run our rendering algorithm. This choice is made based on practical considerations; the V-Cube View algorithm can naturally handle any number of input views.

The algorithm has two main components: i) virtual view depth prediction and ii) Lumi-Net rendering. The whole pipeline is made differentiable and is trained in an end-to-end fashion. 

\begin{figure}[t]
	\centering
	\includegraphics[width=0.99\linewidth]{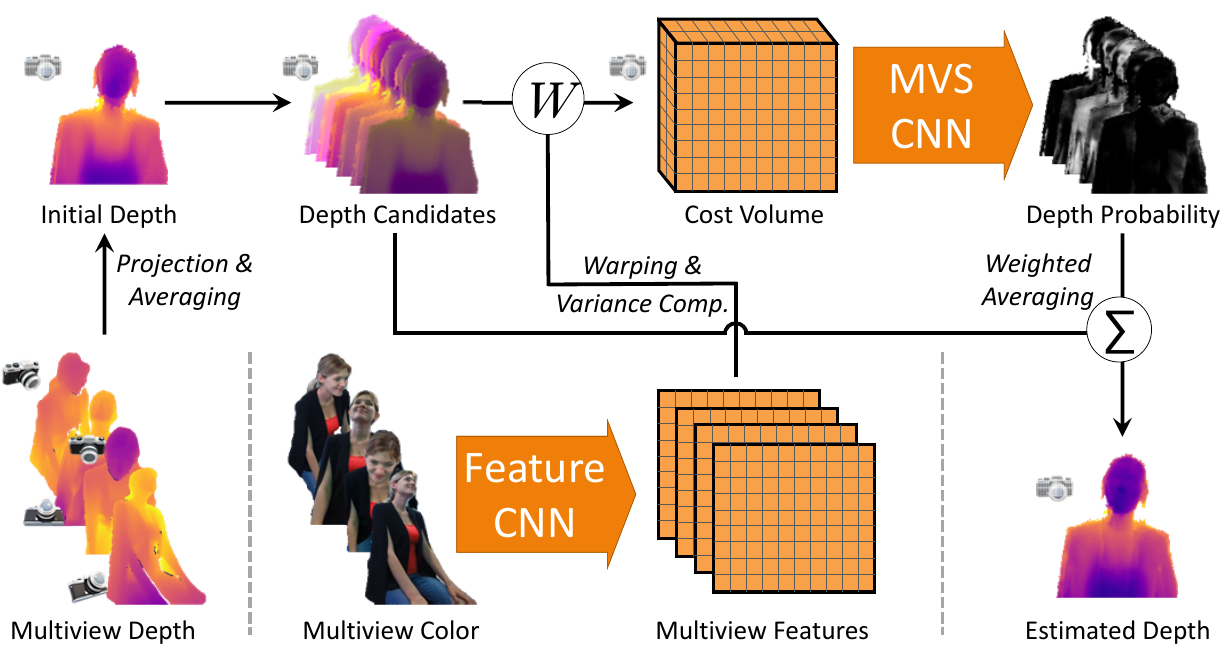}
	\vspace{0pt}
	\caption{Detailed virtual-view depth prediction pipeline.}
	\vspace{0pt}
	\label{fig:3dvideo_pipeline_depth}
\end{figure}

\subsection{Virtual View Depth Prediction}

\begin{figure*}[t]
	\centering
	\includegraphics[width=0.95\linewidth]{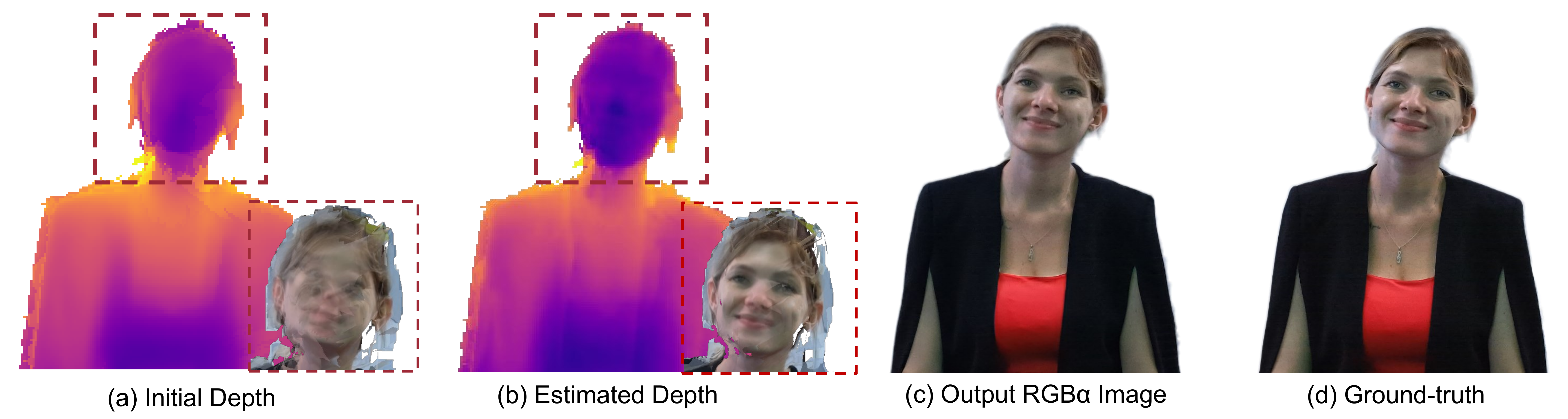}
	\vspace{-8pt}
	\caption{Results of the depth estimation and Lumi-Net rendering modules of the V-Cube View algorithm. (a) The initial virtual-view depth map obtained by fusing the input depth maps. The initial geometry is clearly deficient. In the lower right corner we visualize the quality of the geometry in the face region using a simple texturing scheme that averages the textures of input images. (b) Our depth estimation module significantly improves the geometry. Again, we visualize the quality of the facial region geometry using the same simple texturing scheme as in (a).  (c) This is the RGB$\alpha$ image output by the Lumi-Net rendering module, which compares favorably with the ground-truth shown in (d). The sample is taken from our captured training dataset (validation subset) for illustration.
	}
	\vspace{0pt}
	\label{fig:depth_texture_pred}
\end{figure*}

The first step of our method is to predict a depth map at the given virtual viewpoint, which will be used later to synthesize the texture. For efficiency, the depth prediction module runs at $1/4$ resolution, which we found to be sufficient for our image synthesis. The input contains depth maps measured at four different camera viewpoints, which can be leveraged to determine the virtual view’s depth map. However, these depth measurements are rather noisy. Figure~\ref{fig:depth_texture_pred} (a) shows the depth prediction by warping the four input depth maps to the virtual viewpoint and taking the average of the warped depth values at each pixel. As we can see, the resulting depth map is deficient and it leads to very poor images when texture is applied. 

Our method estimates an accurate virtual-view depth image using multiview stereo matching methodology (MVS) leveraging color images. Thee input depth maps are only used to compute an initial depth map. 
Figure~\ref{fig:3dvideo_pipeline_depth} depicts the detailed algorithm pipeline. 
Let $\{\bI_i,\bD_i\}$ be the input multiview color and depth images. We first use the depth images $\{\bD_i\}$ to derive a set of depth maps at the virtual view, denoted as $\{\bD_i^{'}\}$, which are obtained by projecting the per-view meshes constructed from $\{\bD_i\}$ onto the virtual view image plane followed by rasterization. Then we compute an averaged depth map via $\overbar{\bD} = \frac{\sum_i \bM_i^{'} \cdot \bD_i^{'}}{\sum_i \bM_i^{'}}$, where $\bM_i^{'}$ is the visibility mask of $\bD_i^{'}$ and the calculation is done pixel-wise. This averaged depth is inaccurate and cannot be directly used for image synthesis. Instead, we use $\overbar{\bD}$ as an initial, coarse estimation and apply a convolutional neural network to regress an accurate depth map. Specifically, we define a depth correction range $[-\Delta d, \Delta d]$ and generate $N$ spatially-varying depth hypotheses $\{\widehat{\bD}_k\}$ by uniformly sampling depth corrections $\{\sigma_k\}$ within this range and adding them to the initial depth, i.e., $\widehat{\bD}_k = \overbar{\bD}+\sigma_k$, $k=0,\ldots,N-1$. To construct a cost volume for depth estimation, we apply a 2D CNN on the color images $\{\bI_i\}$  to extract features, denoted as $\{\bF_i\}$. Then we warp the per-view feature map $\bF_i$ to the virtual view with each depth hypothesis $\widehat{\bD}_k$ and compute the feature variance across views, resulting in a cost volume $\bV$ of size $H\times W\times N\times C$, where $C$ is the feature channel. A 3D CNN is applied to $\bV$ to predict a depth probability volume $\bP$ of size $H\times W\times N$. The final depth estimation is computed as $\bD = \sum_{k=1}^N \bP_k \cdot \widehat{\bD}_k$.  

The careful reader will notice that our depth prediction is different from traditional deep MVS depth estimation techniques in computer vision such as \cite{yao2018mvsnet,luo2019p}. We derive this different technique for several reasons. First, we have to predict depth for a virtual view without a color image. Second, we have the initial depth map calculated from the input data and we do not need to use predefined sweeping planes as depth hypotheses. Instead, our depth hypotheses are dynamic and spatially varying with dependence on the initial depth map. In addition, we do not have ground truth depth in our training data and to resolve this issue we apply a weakly-supervised learning scheme that exploits photometric discrepancy of color images for supervision. 

We can visualize the quality of our depth prediction by applying texture to the depth map. Given estimated virtual-view depth $\bD$, we can warp the input-view color images $\{\bI_i\}$ to the virtual view as $\{\bI_i^w =warp(\bI_i|\bD)\}$. The warping function is differentiable with a differentiable mesh rasterizer applied. A naive multiview texture fusion can be obtained by averaging the warped images as $\bI^a = \frac{\sum_i \bM_i^{w} \cdot \bI_i^{w}}{\sum_i \bM_i^{w}}$, where $\bM_i^{w}$ is the visibility mask of $\bI_i^{w}$ and again the calculation is done pixel-wise. As shown in Fig.~\ref{fig:depth_texture_pred}~(a) and (b), the depth map so-obtained is significantly improved when compared with the initial depth, demonstrating the efficacy of our depth prediction module. 

\paragraph{Training loss} To train our method, we capture RGBD images of novel views as the training target (details can be found in Section~\ref{sec:training_data}). Since the captured depth maps are inaccurate, we discard them and never use them in training. In the absence of ground-truth depth, we train the depth prediction module in a weakly-supervised fashion by exploiting photometric discrepancy.  

Concretely, our training losses are the differences between the fused texture $\bI^a$ and the warped multiview images $\{\bI_i^{'}\}$ as well as the target image $\bI^*$: $\mathcal{L}_{color\textrm{-}mv}^d = \sum_i \sum_x \bM_i^{w}(x)\cdot\|\bI^a(x) - \bI_i^{'}(x)\|_1$, $\mathcal{L}_{color\textrm{-}gt}^d = \sum_x \bM(x)\cdot\|\bI^a(x) - \bI^*(x)\|_1$ where $x$ denotes image pixel, $\bM = \cup_i \bM_i^{'}$ is the valid pixel mask of $\bI^a$, and $\|\cdot\|_1$ indicates the $l_1$ norm. We also impose a smoothness prior by adding a second-order smoothness loss $\mathcal{L}_{smooth}^d = \sum_x |\nabla^2 \bD(x)|$ where $\nabla^2$ is the Laplacian operator. Note that since the whole pipeline of our V-Cube View method is differentiable and end-to-end trained, the depth prediction module also receives additional supervision signals from the later modules during training.

\subsection{Lumi-Net Rendering}\label{sec:lumi-net}
A naive way to render is texture averaging, which computes the shading of every surface point by simply averaging this point’s colors among the warped input images in which it is visible. Unfortunately, the warped textures exhibit different levels of details at different regions, depending on their visibility in the original views and their viewpoint difference to the target. Moreover, the target-view depth map cannot be estimated perfectly. The depth error will lead to different degrees of texture distortion in the warped images, especially for our wide-baseline case where viewpoint changes are large. To achieve better texture fusion quality, our Lumi-Net rendering seeks for a viewpoint and geometry-aware texture blending strategy built on two principles of Unstructured Lumigraph Rendering: resolution sensitivity and minimal angular deviation~\cite{buehler2001unstructured}.  We derive a shared CNN for the warped images $\bI_i^w$ to predict the blending weight for each pixel in the image while taking viewpoints and geometry into account.

We start with the resolution sensitivity principle~\cite{buehler2001unstructured}, which is essentially a depth difference prior: textures of object surfaces closer to the camera center are of higher resolution and should be given more weight in blending to ensure sharpness. The depth difference prior is useful for unstructured cameras, especially for our wide-baseline case with large viewpoint differences. Following this principle, we compute for each pixel the depth difference to its corresponding point on an input view given the estimated virtual-view depth map $\bD$. We do this by creating a new virtual-view depth map $\bD_i^{w}$ as follows. For each pixel $p$ in $\bD_i^{w}$, we look up the corresponding 3D surface point using virtual-view depth map $\bD$ and project the resulting point to the input view and obtain its input-view depth value, which is assigned to the pixel $p$. With $\bD_i^{w}$ so computed, we obtain the depth difference as $\Delta\bD_i = \bD_i^{w}-\bD$.

Next, we examine the minimal angular deviation principle~\cite{buehler2001unstructured}, which suggests that the input textures of object surfaces with smaller viewing angle difference to the target view direction should be given greater blending weights. This principle is especially important for non-Lambertian surfaces, such as human faces in video conferencing, because for such surfaces the appearance of a surface point changes significantly with respect to the viewing direction. Following the minimal angular deviation principle, we first compute normalized direction vector maps $\bN$ and $\bN_i^{w}$ based on $\bD$ and $\bD_i^{w}$ respectively. Specifically, for each pixel $p$ of $\bN$, we find the corresponding surface point using depth map $\bD$ and compute the direction vector as the vector from the surface point to the virtual viewpoint. This direction vector is normalized and assigned to pixel $p$ as its value. The computation of $\bN_i^{w}$ is done similarly using $\bD_i^{w}$ and the input viewpoint. 
We then compute the angle map between directions $\bN_i^{w}$ and $\bN$, denoted as $\Delta\bN_i$. 
Adding $\Delta\bN_i$ to the CNN motivates it to increase blending weights for input textures with smaller direction angle deviations.

Besides the warped images and difference maps, we also feed the per-view visibility mask $\bM_i^w$ to the CNN for blending weight prediction. In summary, the input to the CNN is the concatenation of ($\bI_i^w$, $\bM_i^w$, $\Delta\bD_i$, $\Delta\bN_i$) and the output is a blending weight map $\bW_i$. This CNN continues to work on the $4\times$ downsampled image resolution for efficiency. After generating all the weight maps, we apply a pixel-wise \textit{softmax} operator across input views to normalize the blending weights as $\widetilde{\bW}_i = \frac{\exp({\bW_i})}{\sum_j \exp({\bW_j})}$. Then we bilinearly upsample the weight maps by  $4\times$ to the raw resolution, and obtain the blended image as $\bI^b = \sum_i \widetilde{\bW}_i\cdot \bI_i^w$.

Finally, we use a CNN to post-process the blended texture and predict a transparency map. It takes $\bI^b$ as input and produces a four channel output consisting of the final color image $\bI$ and the transparency map $\alpha$. 
The post-processing runs at the full resolution of the image and the CNN has the image processing functions that clean up the blended texture and create the final image. These image processing functions include refining the silhouette boundaries and filling small holes.
We also aim to improve the quality of the face region during post-processing because human faces are particularly important in video conferencing. We do this by adding a special penalty for perceptual loss of the face region. Finally, we improve the overall sharpness of the final image by adding an adversarial learning scheme~\cite{isola2017image} that is widely used for image processing and synthesis, as described below.

\paragraph{Training loss} 
We employ a collection of carefully designed losses to train our Lumi-Net rendering module. The training loss for the blending weight prediction CNN is simple. We add direct supervision to the blended texture $\bI^b$ by minimizing its difference to the ground-truth color image:  $\mathcal{L}_{color}^l = \sum_x\bM(x)\cdot\|\bI^b(x)-\bI^*(x)\|_1$, where $\bM$ is the pixel visibility mask.

The training loss of the post-processing CNN is a bit more complex because of its variety of image processing functions, including refining the silhouette boundaries, filling small holes, improving the face region, and image sharpening.
We first minimize the discrepancy between the output RGB$\alpha$ image and the ground truth by using the loss function $\mathcal{L}_{rgba\textrm{-}gt}^r = \sum_x\|\alpha(x)\cdot\bI(x)-\alpha^*(x)\cdot\bI^*(x)\|_1$. To stabilize the result and avoid overfitting, we also encourage the output image to preserve the input colors via $\mathcal{L}_{color\textrm{-}input}^r = \sum_x\alpha(x)\cdot\bM(x)\cdot\|\bI(x)-\bI^b(x)\|_1$. For better transparency prediction, we add an $\alpha$-map loss as $\mathcal{L}_{\alpha}^r = \sum_x\|\alpha(x)-\alpha^*(x)\|_1$. To improve the fidelity of the face region, we introduce a perceptual loss~\cite{johnson2016perceptual} as
$\mathcal{L}_{face}^r = \sum_l\|\phi_l(crop(\bI))-\phi_l(crop(\bI^*))\|_1$, where $crop(\cdot)$ is the face bounding box cropping operation and $\phi_l(\cdot)$ denotes the multi-layer features from a VGG-19 network~\cite{simonyan2014very} pretrained on ImageNet~\cite{russakovsky2015imagenet}. Finally, for image sharpening we incorporate adversarial learning with a PatchGAN as in \cite{isola2017image} and use the least squares GAN loss from \cite{mao2017least} as $\mathcal{L}_{adv}^r = \|D(\bI)-\mathbf{1}\|^2$, where $D$ is the discriminator network. The adversarial loss for $D$ is $\mathcal{L}_{adv\textrm{-}D}^r = \frac{1}{2}\|D(\bI)-\mathbf{0}\|^2 + \frac{1}{2}\|D(\bI^*)-\mathbf{1}\|^2$.

\subsection{Training Data}\label{sec:training_data}
To capture training data for our method, we setup six RGBD cameras with similar spatial arrangement to the VirtualCube. We place two additional cameras to capture the target-view images for training. 
We add perturbations to the camera poses 
during the data capture campaign 
to introduce view variations. Modest perturbations are added to the six cameras, whereas the target camera poses are flexibly adjusted to cover a wide range of viewpoints. To train the method, we use the left four camera as the input views and synthesize the left target view. Similarly, the right four cameras are used to generate the target view on the right.
During data capture, we ask actors to sit in front of the cameras and make natural actions as in a face-to-face meeting, including face expression, body movements, and some hand gestures. We also captured side-view images by asking actors to sit towards the side screens.

We captured multiview RGBD data of 18 subjects in total. For each subject, multiple video clips are recorded at 15fps. Each subject wears 3-5 personal garments during capture. The dataset contains 920K frames in total, with 51K frames per person on average. The background matting method of \cite{sengupta2020background} is used to obtain alpha map labels of the target view images. We detect face regions by the method of \cite{chen2016supervised}.

\subsection{Temporal Smoothing}

Since our method synthesizes images frame-by-frame, we apply temporal smoothing to the output RGB$\alpha$ sequences to improve temporal consistency especially for boundary regions. Specifically, for the transparency map estimates, we maintain a history buffer $\alpha_h$ and blend the current-frame output $\alpha$ with $\alpha_h$ via $\alpha'=w\alpha+(1-w)\alpha_h$, where scalar $w$ is the blending weight ($w=0.5$ in our implementation). The history buffer is updated at each frame as $\alpha_h=\alpha'$. We apply a similar temporal smoothing strategy for color images, except that we only process border pixels. Let $\bI_h$ be the image history buffer, we first shrink the processed transparency map $\alpha'$ by $n$ pixels ($n=10$ in our implementation) to obtain $\alpha'_{interior}$,  which is achieved by a spatial min-pooling operation with $(2n+1)\times (2n+1)$ kernel size. Let $\alpha'_{border} = \alpha' - \alpha'_{interior}$ be the border region transparency map, we blend the current-frame color image $\bI$ with $\bI_h$ via $\bI' = \big((\alpha'-(1-w)\alpha'_{border})\bI + (1-w)\alpha'_{border}\bI_h\big)/\alpha'$. The history buffer $\bI_h$ is then updated as $\bI_h=w\alpha\bI + (1-w\alpha)\bI_h$. The $(\bI',\alpha')$ pair will be the final RGB$\alpha$ output of our system to be streamed to the remote users.

\section{Experiments}
We implemented three physical VirtualCube instances to evaluate our system in different video communication scenarios. The three VirtualCubes are located in one building and connected in a LAN network with 1Gbps bandwidth. For each VirtualCube, we deploy our software system on a PC with Intel Core i9-10980XE CPU, 64GB memory, and three GPU cards: two Nvidia GeForce RTX 3090 for input RGBD video processing and rendering and one Nvidia GeForce RTX 2080 for displaying the system UI and rendering 
results onto three screens. After the RGBA video frames are captured by six cameras and transmitted to the main memory from USB ports, our system copies all video frames to GPU. The following video segmentation, V-Cube View algorithm and smoothing algorithm are all implemented by Direct3D shaders and executed on the two 3090 GPU cards. For V-Cube View, the neural networks are converted by the NNFusion framework~\cite{rammer-osdi20}, and the 3D warping and rasterization operations are implemented as dedicated HLSL shaders.
Each of the two 3090 GPUs is responsible for rendering the portrait images of the local participant for one view of the remote participants. 
After rendering, the generated frames are copied from GPU to CPU. 
We compress RGB and alpha frames into JPEG images separately and transmit them 
via the TCP/IP protocol.
The transmission bitrate is about 7Mbps.
The receiver decompresses the frames, loads them into the 2080 GPU, composes them with the rendered 3D background, and finally displays them on the  screens. 

Our system achieves real-time performance for meetings between either two or three participants. For a round-table meeting between three participants, the timings of all steps in our system are: 
60ms for RGBD video acquisition, 40ms for rendering a portrait frame on an Nvidia GeForce RTX 3090 GPU, 30ms for copying and compressing the portrait images, 100ms for network transmission, and 40ms for decompressing the portrait images and rendering them on screen. Thus the end-to-end delay between two sites is about 300ms. Since all these steps can be executed in parallel, our system can achieve 23fps for round-table meetings among three participants and 30fps (alternative-frame processing on two GPUs) for one-one meetings between two participants. The viewpoint tracker uses 6 threads to track eye positions and an additional thread for transmission. The viewpoint can be updated at 30HZ and the network delay is 100ms. The overall delay is admissible for our meeting scenario without noticeable discomfort.

Our current system focuses on realistic portrait video rendering to establish visual attention and eye contact. For audio transmission, we use a commercial teleconference software (Microsoft Teams). 
In practice, we found that the video delay of our system is almost the same as the audio transmission delay so no special processing is applied.

\vspace{0pt}
\subsection{VirtualCube Meeting}
We apply our V-Cube system for several video communication scenarios, including face-to-face meetings between two participants, round-table meetings with three participants, and side-by-side meetings between two participants, as shown in Figure~\ref{fig:teaser}. \emph{Please refer to the accompanying video for video clips  recorded live from a camera behind the participant in each meeting setup.}

\vspace{0pt}
\paragraph{Face-to-face meeting} As shown in the accompanying video  ($0'13''-1'10''$), our system faithfully preserves the gaze contact between two participants in a face-to-face meeting. In this video clip, both of the two actors are new subjects not present in the training dataset of our V-Cube View method. For comparison, we also demonstrate the same video conference using the original view of one camera ($1'11''-1'26''$) as done in commercial teleconference software. As the cameras are mounted around the boundary of the large screen to avoid occluding the displayed content, the camera views are far away from the participant's view. We choose the bottom camera which is closest to the participant view in this comparison. It is clear that in this case the mutual eye gaze between participants is not preserved and the participants cannot establish eye contact.

\vspace{0pt}
\paragraph{Round-table meeting} We also evaluate the performance of our system in a round-table meeting among three participants at three different locations. As shown in the accompanying video ($1'32''-3'20''$), our system successfully delivers an immersive meeting experience for all three participants as if they were in the same room. 
Note that when one participant speaks (around $1'50''$), the other two visually pay attention to the speaker and the participants naturally switch their attention to others as the speaker changes. This visual attention is important in real-world meetings for the participants to not only receive vivid visual feedback from others but also achieve smooth turn taking and floor control. 
We also demonstrate that a participant makes side glances and side comments to a remote participant as if they were in the same room  ($2'30''-2'40''$), which cannot be achieved by existing video-based teleconference systems. 
In this video clip, the male actor in yellow appeared in the training dataset of V-Cube View, while the other two actors are unseen. For comparison, we show the same video conference with commercial teleconference software in the supplementary video. Again, the participants are not able to establish eye contact, nor can the visual attention to the speaker be correctly displayed.

\vspace{0pt}
\paragraph{Side-by-side meeting} Thanks to the flexibility of our V-Cube assembly, our system also supports side-by-side meetings between two participants at different locations ($3'26''-4'45'')$, where the working items on the participants' computer screens are visible and shared by both participants. During the meeting, a participant can see whether the remote participant is visually paying attention to a specific work item as desired and naturally switches his attention between his own screen and the other participant, as well as the remote participant's screen as in a real working environment. In this side-by-side meeting video, both of the two actors appeared in the training data of V-Cube View. Note that although  side-by-side meetings are fairly common in real workplaces, there is no existing video conference system that can reproduce this scenario for remote users. 

We have received feedback from dozens of volunteers who have used our system for both face-to-face meetings and round-table group meetings. All these participants felt that our system provides correct eye gaze and eye contact, which leads to a natural and realistic visual communication experience that they have never seen with existing video conferencing software. For round-table group meetings, several participants commented that ``the two remote participants on the screen are in the same room” even though these two participants were actually in different locations. They also enjoyed using our system: ``I love that the system allows one to view their partners via several angles" and ``as though your partners were sitting right across your table".
Some participants commented that the VirtualCube system is “game changing” and “mind blowing”.

\subsection{V-Cube View Evaluation}
In this section, we further evaluate our V-Cube View method, one of the core underpinnings of the VirtualCube system.

\vspace{-1pt}
\paragraph{Quantitative evaluation} For quantitative evaluation, we split our captured dataset of 18 subjects into three sets: \emph{train}, \emph{validation}, and \emph{test}. The training and validation sets contain the same subset of 14 subjects, but differ in the video clips used for each subject. The other 4 subjects are included in the test set as \emph{unseen} persons to verify the generalization of our trained model to new users. These 4 testing subjects have different skin colors (black, brown, white, and yellow) and contain both male and female. We train our method on the training set, and evaluate the results on all three subsets. We report three metrics: PSNR of the generated images, mean square error (MSE) of the alpha maps, and the perceptual error of the synthesized faces (see Section~\ref{sec:lumi-net} for more details regarding the face perceptual error). 

The evaluation results are presented in Table~\ref{tab:quantitative}. As we can see, our method is able to generate high quality images with PSNR as high as $29.2$--$29.8$db. 
It also shows that our trained model generalizes well to unseen persons: the image PSNRs are similar to the train and validation sets, while the alpha map and face perceptual errors are slightly higher. 

We also conduct ablation studies to verify the effectiveness of different components. As shown in Table~\ref{tab:quantitative}, the performance drops significantly if we remove the depth estimation module and simply use the initial depth. Table~\ref{tab:quantitative} also demonstrates the efficacy of our Lumi-Net design: the depth and normal difference priors lead to significantly improved rendering results.

\begin{table}[t!]
	\centering
	\caption{Quantitative evaluation of our V-Cube View method and comparison with different configurations. The \emph{train} set contains video clips of 14 subjects. The \emph{validation} set includes the same 14 subjects but with new video clips which do not appear in the train set. The \emph{test} set contains another 4 subjects unseen during training. Results are averaged on evenly-sampled 19K, 6K, and 6K frames for the three sets, respectively.\label{tab:quantitative}}
	{
		\small
		
		\begin{tabular}{l|ccc|ccc|ccc}
			\hline
			& \multicolumn{3}{c|}{\!\!Image PSNR ($\uparrow$)\!\!}  & \multicolumn{3}{c|}{\!\!\!\!$\alpha$ MSE $\times 10^{-3}$ ($\downarrow$)\!\!\!\!} & \multicolumn{3}{c}{\!\!\!Face per. error ($\downarrow$)\!\!\!}\\
			& \!\!\!train\!\!\! & \!\!\!val.\!\!\! & \!\!\!test\!\!\! & \!\!\!train\!\!\! & \!\!\!val.\!\!\! & \!\!\!test\!\!\! & \!\!\!train\!\!\! & \!\!\!val.\!\!\! & \!\!\!test\!\!\!\\
			\hline
			\!\!\!w/o depth estimation & \!\!\!26.01\!\!\!   & \!\!\!25.73\!\!\!  & \!\!\!27.01\!\!\!   & \!\!\!4.37\!\!\! & \!\!\!4.78\!\!\!\!  & \!\!\!4.60\!\!\!  & \!\!\!0.54\!\!\! & \!\!\!0.54\!\!\!  & \!\!\!0.58\!\!\! \\
			\!\!\!Lumi-Net w/o diff. maps\!\!\!\! & \!\!\!27.84\!\!\!  & \!\!\!27.57\!\!\!  & \!\!\!28.09\!\!\!  & \!\!\!3.41\!\!\! & \!\!\!3.71\!\!\!\!  & \!\!\!4.26\!\!\! & \!\!\!0.50\!\!\! & \!\!\!0.49\!\!\!  & \!\!\!0.54\!\!\!  \\
			\!\!\!Lumi-Net w/o image & \!\!\!28.77\!\!\!   & \!\!\!28.52\!\!\!  & \!\!\!28.90\!\!\!   & \!\!\!2.38\!\!\! & \!\!\!2.60\!\!\!\!  & \!\!\!3.26\!\!\! & \!\!\!0.46\!\!\! & \!\!\!0.45\!\!\!  & \!\!\!0.51\!\!\!  \\
			\!\!\textbf{Our full model} & \!\!\!\!\textbf{29.87}\!\!\!\!   & \!\!\!\!\textbf{29.69}\!\!\!\!  & \!\!\!\!\textbf{29.97}\!\!\!\!   & \!\!\!\!\textbf{1.36}\!\!\!\! & \!\!\!\!\textbf{1.51}\!\!\!\!  & \!\!\!\!\textbf{2.11}\!\!\!\! & \!\!\!\!\textbf{0.39}\!\!\!\! & \!\!\!\!\textbf{0.38}\!\!\!\!  & \!\!\!\!\textbf{0.45}\!\!\!\! \\
			\hline
		\end{tabular}
		\vspace{-2pt}
	}
\end{table}

\begin{figure}[t!]
    \centering
    \includegraphics[width=\columnwidth]{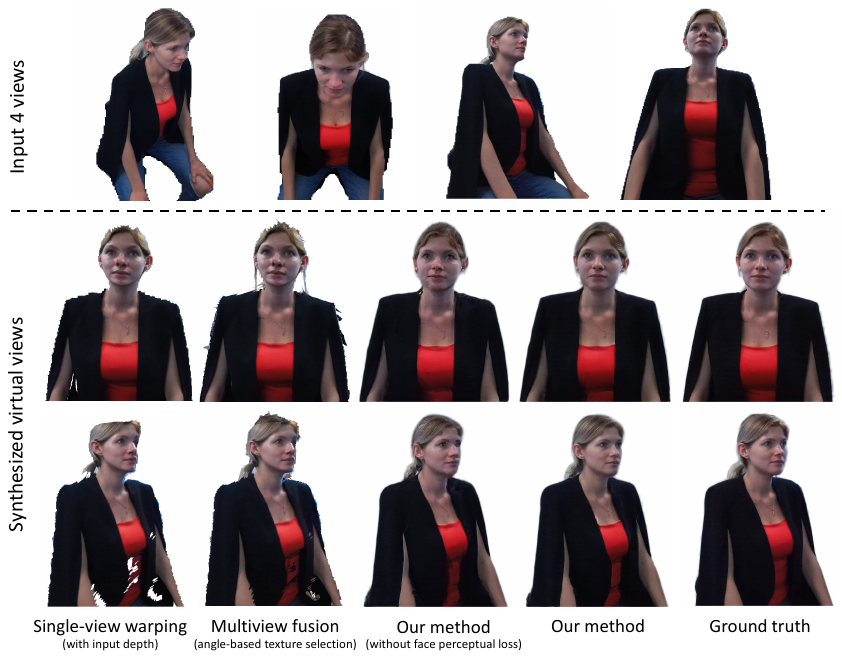}
    \vspace{-18pt}
    \caption{Rendering results of our method compared to other alternatives. See text for details. 
     (\textbf{Best viewed on screen with zoom-in})}
    \label{fig:results_compare}
\end{figure}

\paragraph{Qualitative evaluation} 
In Fig.~\ref{fig:results_compare} we show some sample input and output from our method. We also show the results from other alternative solutions, including single-view image warping (best input view used here) and a multiview depth and texture fusion scheme similar to \cite{tola2009virtual}. Visually inspected, our method produces realistic rendering results significantly better than the compared solutions. We also present the results of our method trained without the perceptual loss on face region, where the degraded face fidelity can be clearly observed.

We also asked randomly selected users of our system to visually inspect the rendered still images and video clips randomly drawn from our validation and test sets and compare them against the ground truth. Most of them commented that our rendering results look “very realistic” and they could not easily distinguish them from the ground truth images.

\vspace{-10pt}
\paragraph{Running time} The V-Cube View method takes about 40ms to render one image. The depth estimation module and blending weight prediction CNN take 5ms and 2ms, respectively. The post-processing CNN takes about 20ms, which is the most time-consuming component as it runs at $1280\times 960$ resolution. We have tried running all components at $1280\times 960$ resolution, with which the whole method is $5\times$ slower. The warping operations and temporal smoothing are all extremely fast ($\sim$ 1ms), thanks to our HLSL shader implementation. Other parts such as IO between CPU and GPU takes about 9ms.

\section{Conclusion and Future Work}\label{sec:discussions}

The VirtualCube system advances the state-of-art of large-format video conferencing and shows early promise of the possibility of bringing remote parties together and letting them interact as if they were in the same room. Furthermore, we show that VirtualCube can be used as the basic building blocks of video conferences. This is a viable and versatile way to model video conferences – indeed this is a new way to think about video conferencing as well. We believe there is much more to be discovered along this direction. The fact that the VirtualCube system can be built completely with 
off-the-shelf hardware will inspire more people to join this research direction and build the next generation video conferencing technologies on top of our work.  

One possible future work is to make VirtualCube more flexible. In this work we assume that each VirtualCube is used by one meeting participant. This assumption is made mainly for simplification and for the fact that this is one of the most common application scenarios. There is no fundamental reason to prevent more users from sharing a VirtualCube.  
Incorporating spatial audio into our system will also enhance the immersive meeting experience. Our current work focuses only on the visual attention aspect of video conferencing, and a commercial software is used for audio recording and transmission.  Much excellent research has been done in spatial audio \change{\cite{garner2017echoes,kim2019immersive}} and we leave it as future work. Another promising avenue for future research is to increase the ability to support complicated hand gestures. The VirtualCube currently supports common simple gestures. Handling arbitrary hand gestures is a challenging task. In particular, the Azure Kinect camera we use cannot acquire high-quality depth data for moving hands. In addition, RGB videos of fast hand gestures can be blurry. For these reasons, advances in dealing with hand gestures will require both progress in camera hardware and innovations in software algorithms.

\bibliographystyle{abbrv-doi}

\bibliography{vc}
\end{document}